\documentclass[preprint,5p]{elsarticle}

\usepackage{booktabs}
\usepackage{multirow}
\usepackage{subfigure}
\usepackage{url}

\journal{arXiv}

\begin{document}

\begin{frontmatter}

\title{FEAFA+: An Extended Well-Annotated Dataset for Facial Expression Analysis and 3D Facial Animation}

\author[1]{Wei Gan}
\ead{ganwei19@mails.ucas.ac.cn}

\author[1]{Jian Xue}
\ead{xuejian@ucas.ac.cn}

\author[1]{Ke Lu\corref{cor}}
\ead{luk@ucas.ac.cn}

\author[1]{Yanfu Yan}
\ead{yanyanfu16@mails.ucas.ac.cn}

\author[1]{Pengcheng Gao}
\ead{gaopengcheng15@mails.ucas.ac.cn}

\author[1]{Jiayi Lyu}
\ead{lyujiayi\_cnu@163.com}

\cortext[cor]{Corresponding author}
\address[1]{School of Engineering Science, University of Chinese Academy of Sciences, Beijing 100049, China}

\begin{abstract}
  Nearly all existing Facial Action Coding System-based datasets that include facial action unit (AU) intensity information annotate the intensity values hierarchically using A--E levels. However, facial expressions change continuously and shift smoothly from one state to another. 
  Therefore, it is more effective to regress the intensity value of local facial AUs to represent whole facial expression changes, particularly in the fields of expression transfer and facial animation.
  We introduce an extension of FEAFA in combination with the relabeled DISFA database, which is available at \url{https://www.iiplab.net/feafa+/} now. 
  Extended FEAFA (FEAFA+) includes 150 video sequences from FEAFA and DISFA, with a total of 230,184 frames being manually annotated on floating-point intensity value of 24 redefined AUs using the Expression Quantitative Tool.
  We also list crude numerical results for posed and spontaneous subsets and provide a baseline comparison  for the AU intensity regression task.
\end{abstract}


\end{frontmatter}

\section{Introduction}
\label{sec:introduction}

Human emotional expression is multimodal, and facial expression plays the most important role in expressing emotions. Driven by the rapid development of computer vision and deep learning technology, automated facial expression recognition and analysis has become a popular research direction and can be applied to many fields such as human-computer interaction, driver fatigue detection, virtual avatars, and patient pain estimation.
There are two common approaches to describe expressions: the judgement-based approach classifies expressions into six prototypical categories based on Ekman's Basic Emotion Theory, and the sign-based approach combines coded facial muscle actions such as action units (AUs) to explain expressions \cite{harrigan2008new}.

The Facial Action Coding System (FACS) proposed by Ekman et al. \cite{friesen1978facial} in 1978 and revised in 2002 is the most comprehensive and widely used facial expression description system.
FACS 1978 distinguishes 44 unique facial AUs and action descriptors (ADs) on the basis of the types and motion characteristics of facial muscles.
FACS 2002 \cite{Ekman2002Facial} is the current and only specified version for reference, and includes 27 facial AUs, 25 codes of head and eye positions, and 28 codes of miscellaneous movements. 
The 2002 revision also allows for a 5-level (A--E) ordinal scale of AU intensity, which quantitatively interprets the intensity from a trace to the maximum strength of an action.
With the help of the FACS, various facial expressions can be decomposed to combinations of AUs. 

With the prevalence of the FACS, automated AU detection and AU intensity estimation technology has been widely concerned and applied. 
The automatic detection and estimation of AU groups by computers plays a big role in accurately analyzing facial expressions and individual emotions. 
As a result of the constant refinement of the technologies of face detection, alignment and recognition, AU detection and intensity estimation have made significant progress and encountered many challenges simultaneously.
A challenges is the insufficiency of well-annotated data \cite{LiYong2020}.
Establishing FACS databases requires specially trained experts and is time-consuming and costly.
Individual facial expressions vary from person to person, and the combined effects of different facial features differ in intensity.

In the early years, research institutions tended to record videos and images of posed facial expressions in the laboratory environment under certain conditions, and produced several commonly used datasets such as CK \cite{kanade2000comprehensive}, MMI \cite{pantic2005web} and Bosphorus \cite{Aly20083D}.
As spontaneous facial expressions started to gain attention, datasets like UNBC-McMaster \cite{Lucey2011Painful}, SEMAINE \cite{Mckeown2012The}, DISFA \cite{Mavadati2013DISFA} and BP4D \cite{Zhang2013A}. 
As computer performance and the development of machine learning are upgraded, larger amounts of data and richer individual expression are required in the field of facial expression analysis.
Data collected from the open environment and the Internet can satisfy demand, and then comes the datasets, for example, AM-FED \cite{Mcduff2013Affectiva}, EmotioNet \cite{fabian2016emotionet}.

However, the FACS-based annotations of the early datasets primarily focused on AU occurrence. 
A few of the later annotations adopted the labels of 5-level AU intensity.
Considering the continuity of expression changes, FEAFA \cite{yan2019feafa} first attempted to explain AUs in a continuous manner based on the FACS and code the intensity using a floating point number.

\begin{figure}[t]
	\centering
	\includegraphics[width=0.9\columnwidth]{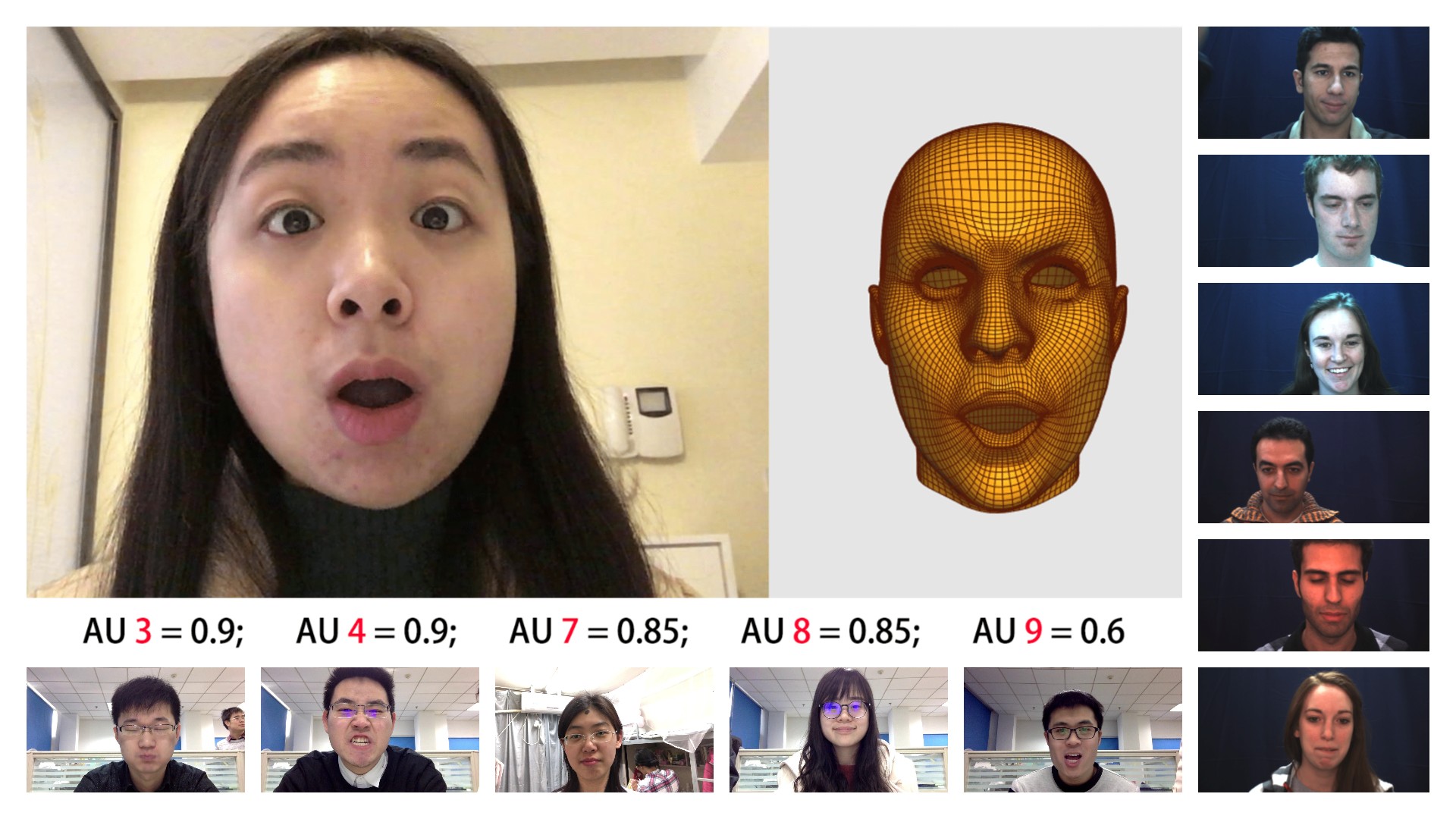}
	\caption{Sample images in the FEAFA+ dataset. The big picture in the upper left corner presents one facial image with its corresponding annotation and expression blendshape. The small pictures in the bottom row and left column are from FEAFA and relabeled DISFA respectively.}
	\label{fig:sample}
\end{figure}

To enrich the data in FEAFA, which only contains posed sequences, we added the available spontaneous sequences from DISFA and annotated the images with 24 defined AUs in the form of floating point numbers, which are available now at \url{https://www.iiplab.net/feafa+/}. We call the combined dataset FEAFA+: An Extended Well-Annotated Dataset for Facial Expression Analysis and 3D Facial Animation. Sample images from FEAFA+ are shown in Fig \ref{fig:sample}. FEAFA+ contains the following:
\begin{itemize}
	\item {\verb|Facial videos|}: 127 available posed sequences from FEAFA of 122 subjects recorded in the wild, and 27 spontaneous sequences from DISFA of 27 subjects recorded in the laboratory which cannot be provided directly by us according to the agreement on the use of DISFA video data.
	
	\item {\verb|Labeled frames|}: 230,184 frames manually annotated using the Expression Quantitative Tool (ExpQuantTool) on the intensities with the 0-1 floating--point representation of 24 defined AUs (i.e. nine symmetrical and 10 asymmetrical FACS AUs, two symmetrical and two asymmetrical FACS ADs). However, there are 9,690 images from DISFA may not be provided directly by us according to the use agreement.
	
	\item {\verb|Baseline regression|}: Baseline performance of a joint AU intensity regression algorithm with CNN architectures on this dataset and and result comparison of baseline regressors.
\end{itemize}

In the remainder of this paper, in Section \ref{sec:background}, we review and summarize several related existing datasets; in Section \ref{sec:feafa+}, we describe the annotation details and numerical analysis of FEAFA+; in Section \ref{sec:baseline}, we present a regression baseline for AU intensity estimation; and in Section \ref{sec:conclusions}, we conclude the paper.


\section{Related Work}
\label{sec:background}

\begin{table*}[htb]
	\centering
	\caption{Public facial expression datasets with FACS annotations in the last decade.}
	\label{tab:datasets_compare}
	\setlength{\tabcolsep}{2pt}
	\scriptsize
	\begin{tabular}{l|l|l|l|l|l}
		\toprule
		\textbf{Dataset} & \textbf{Year} & \textbf{Dataset Information} & \textbf{Data Source} & \textbf{FACS AU related} & \textbf{Posed vs Spontaneous}  \\ \midrule
		\multirow{2}{*}{CK+ \cite{Lucey2010The}}                                                      & \multirow{2}{*}{2010} & 123 subjects / 593 videos / 10,708 frames       & \multirow{2}{*}{In the lab}                                                                & \multirow{2}{*}{\begin{tabular}[c]{@{}l@{}}1, 2, 4, 5, 6, 9, 12, 15, 17, 20, \\ 25, 26\end{tabular}}                            & \multirow{2}{*}{Posed \& Spontaneous} \\ \cline{3-3}
		&                       & Occurrence(subset) \& Intensity (peak frames)  &                                                                                            &                                                                                                                                 &                                       \\ \hline
		\multirow{2}{*}{\begin{tabular}[c]{@{}l@{}}UNBC-\\ McMaster \cite{Lucey2011Painful}\end{tabular}} & \multirow{2}{*}{2011} & 129 subjects /   200 videos / 48,398 fames      & \multirow{2}{*}{In the lab}                                                                & \multirow{2}{*}{\begin{tabular}[c]{@{}l@{}}4, 6, 7, 9, 10, 12, 20, 25, 26, 27, \\ 43\end{tabular}}                              & \multirow{2}{*}{Spontaneous}          \\ \cline{3-3}
		&                       & Occurrence \& 5-level Intensity                &                                                                                            &                                                                                                                                 &                                       \\ \hline
		\multirow{2}{*}{AM-FED \cite{Mcduff2013Affectiva}}                                                   & \multirow{2}{*}{2013} & 242 subjects /   242 videos / 168,359 frames    & \multirow{2}{*}{In the wild}                                                               & \multirow{2}{*}{2, 4, 5, 9, 12, 14, 15, 17, 18, 26}                                                                             & \multirow{2}{*}{Spontaneous}          \\ \cline{3-3}
		&                       & Occurrence only                                &                                                                                            &                                                                                                                                 &                                       \\ \hline
		\multirow{2}{*}{DISFA \cite{Mavadati2013DISFA}}                                                    & \multirow{2}{*}{2013} & 27 subjects /   54 videos / 130,000 frames     & \multirow{2}{*}{In the lab}                                                                & \multirow{2}{*}{\begin{tabular}[c]{@{}l@{}}1, 2, 4, 5, 6, 9, 12, 15, 17, 20,\\ 25, 26\end{tabular}}                             & \multirow{2}{*}{Spontaneous}          \\ \cline{3-3}
		&                       & 5-level Intensity                             &                                                                                            &                                                                                                                                 &                                       \\ \hline
		\multirow{2}{*}{\begin{tabular}[c]{@{}l@{}}MMSE \cite{zhang2016multimodal}\\ (BP4D+)\end{tabular}}   & \multirow{2}{*}{2016} & 140 subjects / 560 videos / 1,400,000 frames & \multirow{2}{*}{In the lab}                                                                & \multirow{2}{*}{\begin{tabular}[c]{@{}l@{}}1, 2, 4, 5, 6, 7, 9, 10, 11, 12, 13, \\ 14, 15, 16, 17, 18, 19\end{tabular}}         & \multirow{2}{*}{Spontaneous}          \\ \cline{3-3}
		&                       & Occurrence \& 5-level Intensity (parts)        &                                                                                            &                                                                                                                                 &                                       \\ \hline
		\multirow{2}{*}{EmotioNet \cite{fabian2016emotionet}}                                                & \multirow{2}{*}{2016} & 1,000,000 frames      & \multirow{2}{*}{\begin{tabular}[c]{@{}l@{}}In the wild\\ (From the Internet)\end{tabular}} & \multirow{2}{*}{\begin{tabular}[c]{@{}l@{}}1, 2, 4, 5, 6, 9, 12, 17, 20, 25, \\ 26, 43\end{tabular}}                            & \multirow{2}{*}{--}                   \\ \cline{3-3}
		&                       & Intensity                                      &                                                                                            &                                                                                                                                 &                                       \\ \hline
		\multirow{2}{*}{GFT \cite{girard2017sayette}}                                                      & \multirow{2}{*}{2017} & 96 subjects /   96 video / 172,800 frames  & \multirow{2}{*}{In the lab}                                                                & \multirow{2}{*}{\begin{tabular}[c]{@{}l@{}}1, 2, 4, 5, 6, 7, 9, 10, 11, 12, 14, \\ 15, 17, 18, 19, 22, 23, 24, 28\end{tabular}} & \multirow{2}{*}{Spontaneous}          \\ \cline{3-3}
		&                       & Occurrence \& 5-level Intensity(parts)         &                                                                                            &                                                                                                                                 &                                       \\ \hline
		\multirow{2}{*}{FEAFA \cite{yan2019feafa}}                                                    & \multirow{2}{*}{2019} & 122 subjects /   123 videos / 99,356 frames     & \multirow{2}{*}{In the wild}                                                               & \multirow{2}{*}{\begin{tabular}[c]{@{}l@{}}2, 4, 5, 9, 10, 12, 16, 17, 18, 20, \\ 28, 29, 26, 30, 34, 43\end{tabular}}          & \multirow{2}{*}{Posed}                \\ \cline{3-3}
		&                       & {[}0-1{]} floating point intensity             &                                                                                            &                                                                                                                                 &                                       \\ \hline
		\multirow{2}{*}{Aff-Wild2 \cite{kollias2019expression}}                                                & \multirow{2}{*}{2019} & 62 subjects /   63 videos / 397,800 frames      & \multirow{2}{*}{\begin{tabular}[c]{@{}l@{}}In the wild\\ (From the Internet)\end{tabular}} & \multirow{2}{*}{1, 2, 4, 6, 12, 15, 20, 25}                                                                                     & \multirow{2}{*}{Spontaneous}          \\ \cline{3-3}
		&                       & Occurrence only                                &                                                                                            &                                                                                                                                 &                                       \\ \hline
		\multirow{2}{*}{FEAFA+}                                                   & \multirow{2}{*}{2021} & 149 subjects /   150 videos / 230,163 frames    & \multirow{2}{*}{In the wild \& lab}                                                        & \multirow{2}{*}{\begin{tabular}[c]{@{}l@{}}2, 4, 5, 9, 10, 12, 16, 17, 18, 20, \\ 28, 29, 26, 30, 34, 43\end{tabular}}          & \multirow{2}{*}{Posed \& Spontaneous} \\ \cline{3-3}
		&                       & {[}0-1{]} floating point intensity             &                                                                                            &                                                                                                                                 &                                       \\  \bottomrule
	\end{tabular}
\end{table*}

Early datasets based on the FACS, such as CK \cite{kanade2000comprehensive}, MMI \cite{pantic2005web}, and Bosphorus \cite{Aly20083D}, were mostly collected in constrained environments and required the subjects to make different faces in a fixed head posture. 
Thus, the video and image data were often in a uniform format and of high quality with rich and varied posed expressions.
The subjects in MMI datasets \cite{pantic2005web} were asked to make posed expressions and minimize their head motions. 
AU occurrences were fully coded, and AU intensities that indicated neutral, onset, apex and offset phases were partially annotated at the frame level.


However, compared with posed facial expressions, spontaneous facial behaviors are less intense and more common in daily life, and have received increasing attention. 
Since 2010, increasing numbers of institutions have shown interest in individual spontaneous expressions, and induced the subjects' expressions in restricted conditions using video clips, interactions and stimulus tasks. 
The expressions produced are greatly influenced by the manner of induction.

CK+ \cite{Lucey2010The} is an extension of the Chon-Kanade dataset,  and consists of 593 sequences and 10,708 frames of posed and spontaneous (smile) facial expressions from 123 subjects. 
The UNBC-McMaster database \cite{Lucey2011Painful} contains 129 subjects with shoulder pain recorded in a laboratory room, from which 200 video sequences of spontaneous facial expressions were produced. 
The SEMAINE database \cite{Mckeown2012The} contains large audiovisual data obtained through 959 conversations between 150 participants and individual Sensitive Artificial Listener (SAL). 
The BP4D database \cite{Zhang2013A} contains data from eight tasks designed to induce 41 young adults to make natural facial expressions.
The GFT database \cite{girard2017sayette} was the first to be published with open and well-annotated facial expression data of multiple people's natural interactions, from 32 recorded three-person groups of 96 subjects.

The DISFA dataset \cite{Mavadati2013DISFA} contains 53 facial expression sequences recorded from 27 young adults while they viewed video clips in a laboratory with a stereo-vision system and uniform illumination. 
Approximately 130,000 frames were manually FACS coded based on occurrence (presence/absence) and intensity (0--5 level) for 12 AUs. 
The extended DISFA dataset (DISFA+) \cite{mavadati2016extended} additionally captures the posed facial expressions of nine participants under guidance. 


After 2013, most of the datasets released were collected in open scenarios such as AM-FED \cite{Mcduff2013Affectiva} or even from the internet such as EmotioNet \cite{fabian2016emotionet}. 
Relaxing restrictions on collection conditions makes it easier to obtain abundant video and image data with various individual expressions.
However, problems remain, such as portraiture rights and inaccurate labeling of separatated images.

The AM-FED dataset \cite{Mcduff2013Affectiva} contains 242 online participants' facial expressions recorded using webcams while they were watching amusing commercials, and 168,359 FACS-coded images for the occurrences of 10 symmetrical AUs and four unilateral AUs. 
The EmotioNet database \cite{fabian2016emotionet} contains one million facial expression images downloaded from the internet, among which 90\% were automatically annotated for AU occurrence and intensity using EmotioNet trained on  existing datasets and 10\% were manually FACS coded to test the detection accuracy.
The Aff-Wild2 database \cite{kollias2019expression} is an extension of the original Aff-Wild dataset \cite{kollias2019deep} that includes downloaded web videos and partially annotated basic expressions and AUs. The AU subset includes 63 videos (397,800 frames with AU occurrences coded) of 62 subjects.
The RAF-AU database \cite{yan2020raf} is an updated version of the RAF-ML database \cite{li2019blended} that consists of 4,601 images from the latter database that were also downloaded from the internet and annotated  with the occurrences of 26 AUs.


The datasets mentioned above mostly provide data with labels of AU occurrence and may also provide 5-level intensity based on FACS.
As facial expressions change smoothly rather than step by step, attempts have also been made to explain AU intensity continuously.
The FEAFA dataset \cite{yan2019feafa} provides continuously numerical labels of AU intensity based on FACS and contains 123 posed videos (99,356 frames) of 122 subjects recorded in real-world conditions under a specific protocol. We extend the FEAFA dataset both in terms of size and video type by relabeling the DISFA dataset \cite{mavadati2012automatic, Mavadati2013DISFA} in the same manner as FEAFA and call the resulting dataset FEAFA+.

Facial expression datasets with FACS annotations from the last 10 years are summarized in Table \ref{tab:datasets_compare}.


\section{FEAFA+ Dataset}
\label{sec:feafa+}

\subsection{Profile of FEAFA+}

The FEAFA dataset contains 123 videos of 122 subjects recorded using an ordinary monocular camera in the wild, and 99,356 frames are manually labeled with (0-1) floating point numbers on 24 AUs chosen from the FACS.
All the participants were asked to make the required facial movements in the Facial Expression Data Recording and Labeling Protocol; thus, the FEAFA dataset only contains posed facial expression data, which cannot fully meet the demands of scientific study.

The original DISFA dataset, published in 2013, consists of the spontaneous expressions of 27 subjects in the laboratory who were asked to watch selected video clips.
The final available dataset contains 27 videos, and each video has 4,845 facial expression images. They coded the AU intensity of each video frame using a 5-level ordinal scale.

We applied for permission to use the DISFA and FEAFA datasets and relabeled the facial images of DISFA in the same manner as FEAFA. 
We relabeled a total of 27 videos (SN001, SN002, SN003, SN004, SN005, SN006, SN007, SN008, SN009, SN010, SN011, SN012, SN013, SN016, SN017, SN018, SN021, SN023, SN024, SN025, SN026, SN027, SN028, SN029, SN030, SN031, and SN032) using a (0--1) floating-point numbers rather than a 6-point intensity scale.
However, for SN005 and SN009 we may only provide labels without images according to the agreement on the use of DISFA.

FEAFA+ is composed of the FEAFA dataset and the relabeled DISFA dataset. 
FEAFA is a posed facial action intensity dataset whose data source comes from multi-generational Asian subjects recorded in varied environments, whereas the relabeled DISFA is a spontaneous dataset composed of data recorded using a specific stereo-vision system in a specified environment and the participants were mainly young adults of various races. 
Thus FEAFA+ can provide much more abundant facial action data than FEAFA, not only becuase of the larger amount of data, but also because FEAFA+ contains both posed and spontaneous facial expression changes of multi-ethnic and multi-generational subjects.

\subsection{AU Annotation}

For FEAFA, nine symmetrical AUs, 10 asymmetrical (unilateral) AUs, two symmetrical ADs and two asymmetrical ADs were chosen and reorganized based on the definitions in the FACS; most facial expressions can be described in this manner. The symmetrical AUs (e.g., Eye Closure) were separated into left and right parts and some asymmetrical AUs of the mouth region (e.g., Lip Suck) were split into upper and lower parts. The definitions of the 24 selected AUs and their corresponding FACS AU/AD are listed in Table \ref{tab:aulist}.

\begin{table*}[htb]
	\centering
	\caption{Explanation of redefined AUs from FEAFA \cite{yan2019feafa}. AU43 (Eye Closure) in FACS is split into two different AUs as Left and Right Eye Close, in the same manner as AU2, AU4, AU5, AU20, and AU30 in FACS. Additionally, AU28 (Lip Suck) in the FACS is subdivided into Upper and Lower Lip Suck.}
	\label{tab:aulist}
	\setlength{\tabcolsep}{1.8mm}
	\footnotesize
	\begin{tabular}{c|l|l||c|l|l}
		\toprule
		\textbf{AU} &  \textbf{Our Definition} & \textbf{FACS No. and Definition} &  \textbf{AU} &  \textbf{Our Definition} & \textbf{FACS No. and Definition} \\ \midrule
		1 &Left Eye Close	   &AU43{ }{ }Eye Closure		  &13&Right Lip Corner Pull	  &AU12{ }{ }Lip Corner Puller\\ \hline
		2 &Right Eye Close 	   &AU43{ }{ }Eye Closure		  &14&Left Lip Corner Stretch &AU20{ }{ }Lip stretcher\\ \hline
		3 &Left Lid Raise 	   &AU 5{ }{ }{ }Upper lid raiser &15&Right Lip Corner Stretch&AU20{ }{ }Lip strecher\\ \hline
		4 &Right Lid Raise 	   &AU 5{ }{ }{ }Upper lid raiser &16&Upper Lip Suck 		  &AU28{ }{ }Lip Suck\\ \hline
		5 &Left Brow Lower 	   &AU 4{ }{ }{ }Brow lowerer	  &17&Lower Lip Suck		  &AU28{ }{ }Lip Suck\\ \hline
		6 &Right Brow Lower    &AU 4{ }{ }{ }Brow lowerer	  &18&Jaw Thrust			  &AD29{ }{ }Jaw Thrust\\ \hline
		7 &Left Brow Raise 	   &AU 2{ }{ }{ }Outer brow raiser&19&Upper Lip Raise		  &AU10{ }{ }Upper Lip Raiser\\ \hline
		8 &Right Brow Raise    &AU 2{ }{ }{ }Outer brow raiser&20&Lower Lip Depress		  &AU16{ }{ }Lower Lip Depressor\\ \hline
		9 &Jaw Drop 		   &AU26{ }{ }Jaw Drop			  &21&Chin Raise  			  &AU17{ }{ }Chin Raiser\\ \hline
		10&Jaw Slide Left      &AD30{ }{ }Jaw Sideways		  &22&Lip Pucker			  &AU18{ }{ }Lip Pucker\\ \hline
		11&Jaw Slide Right 	   &AD30{ }{ }Jaw Sideways		  &23&Cheeks Puff			  &AD34{ }{ }Puff\\ \hline
		12&Left Lip Corner Pull&AU12{ }{ }Lip Corner Puller	  &24&Nose Wrinkle			  &AU 9{ }{ }{ }Nose wrinkler\\ \bottomrule
	\end{tabular}
\end{table*}

In pursuit of more precise explanations of AU intensity than five levels, FEAFA coded each AU using a floating-point number ranging from 0 to 1 with two decimal places. We followed the annotation rules as FEAFA and used the ExpreQuantTool, which directly visualize the combinations of AUs, to improve efficiency and accuracy.

We trained 20 coders and assigned three coders to each video.
After acquiring the available DISFA video sequences, the coders relabeled the videos manually frame by frame. 
Finally, a FACS coding expert checked the labeling results and modified obviously invalid values. For one video, we took the average of the group of labels to reduce inter-observer differences.


After the labeling task was complete, we needed know whether the annotation results were valid and reliable. 
From the definitions of various types of reliability, interrater reliability, which indicates the variation between two or more raters measuring the same subject group, was most suitable for measuring the error of multi-person labeling to validate our annotation \cite{koo2016guideline}.
And the intraclass correlation coefficient (ICC) is widely used in reliability analyses as one of several reliability indices. 

To evaluate annotation reliability, four coders from 10 annotated four randomly chosen videos from the original FEAFA dataset. 
Among the 10 forms of ICC defined by McGraw and Wong \cite{mcgraw1996forming}, we choose the two-way mixed-effects model with the definition of absolute agreement and the type of multiple raters, ICC (2, k), to perform calculations for interobserver reliability analysis. 
The formula to calculate ICC (2, k) for one AU is:
\begin{equation}
	\label{eq:ICC(2,k)}
	\frac{MS_{R}-MS_{E}}{MS_{R}+\frac{MS_{C}-MS_{E}}{n}}
\end{equation}
where $n$ is the number of subjects, $k$ is the number of raters or measurements, $MS_{R}$ is the mean square of the rows and each row represents the intensity values of one frame given by different raters, $MS_{C}$ is the mean square of the columns and each column represents the intensity values of different frames given by one rater, and $MS_{E}$ is the mean square of the error. 

The ICC values vary in the range [0,1], where 0 represents no reliability and 1 represents meaning perfect reliability among the raters. 
Generally, the 95\% confident interval (CI) of the ICC estimate rather than ICC estimate itself should be used to evaluate the level of reliability. A value less than 0.5 is considered to be poor, between 0.5 and 0.75 is considered to be moderate, between 0.75 and 0.9 is considered to be good, and greater than 0.9 is considered to be excellent. In different application fields, the boundaries may vary accordingly.


We used the coding results of four annotators on four videos (12,403 frames in total per coder) in the calculation of inter-observer reliability. 
As shown in Table \ref{tab:ICC_result}, the ICC estimates and their 95\% CIs for 24 AUs are reported based on a mean-rating, absolute-agreement, two-way mixed-effects model ICC (2, k) ($k=4$). 

The results demonstrated that the interrater reliability of most AUs was good or excellent for those whose lower bounds of the 95\% CIs were greater than 0.75, or even 0.9. However, the ICC estimates of AU14 and AU15, which were redefined as Left and Right Lip Corner Stretch, were relatively low, which verifies that the annotation of these AUs needed to be more specific to reduce the subjective judgment bias among the raters.

\begin{table*}[htb]
	\centering
	\caption{Inter-observer reliability of four annotators on four videos(12,403 frames in total).}
	\label{tab:ICC_result}
	\footnotesize
	\begin{tabular}{c|c|c||c|c|c}
		\toprule
		\textbf{AU} & \textbf{ICC estimate} & \textbf{$95\%$ CI} & \textbf{AU} & \textbf{ICC estimate} & \textbf{$95\%$ CI} \\ \midrule
		1 & 0.969 & $0.966-0.971$ &13& 0.927 & $0.924-0.929$ \\ \hline
		2 & 0.969 & $0.966-0.972$ &14& 0.739 & $0.731-0.747$ \\ \hline
		3 & 0.964 & $0.963-0.965$ &15& 0.769 & $0.761-0.776$ \\ \hline
		4 & 0.964 & $0.963-0.965$ &16& 0.930 & $0.927-0.934$ \\ \hline
		5 & 0.936 & $0.933-0.939$ &17& 0.946 & $0.942-0.949$ \\ \hline
		6 & 0.935 & $0.931-0.938$ &18& 0.865 & $0.861-0.869$ \\ \hline
		7 & 0.962 & $0.961-0.964$ &19& 0.960 & $0.958-0.963$ \\ \hline
		8 & 0.963 & $0.962-0.964$ &20& 0.966 & $0.965-0.967$ \\ \hline 
		9 & 0.991 & $0.990-0.992$ &21& 0.889 & $0.884-0.894$ \\ \hline
		10& 0.925 & $0.923-0.928$ &22& 0.888 & $0.879-0.896$ \\ \hline
		11& 0.921 & $0.918-0.923$ &23& 0.992 & $0.992-0.992$ \\ \hline
		12& 0.917 & $0.914-0.920$ &24& 0.828 & $0.822-0.834$ \\ \bottomrule
	\end{tabular}
\end{table*}

\subsection{Numerical Analysis}

FEAFA+ contains 150 videos and 230,163 images with labels, of which 99,356 frames were posed and taken from FEAFA and 130,807 frames were spontaneous and taken from relabeled DISFA. 
Posed and spontaneous facial expressions often involve different facial muscles, which means that, even for the same expression, the AU combination of the posed and spontaneous facial expressions may differ. 
Additionally, we usually pay more attention to spontaneous facial expression changes in daily life. 
We compared the occurrence and distribution of each AU between FEAFA and relabeled DISFA, and also the correlation between each AU within each dataset.

\begin{figure}[t]
	\centering
	\includegraphics[width=0.95\columnwidth]{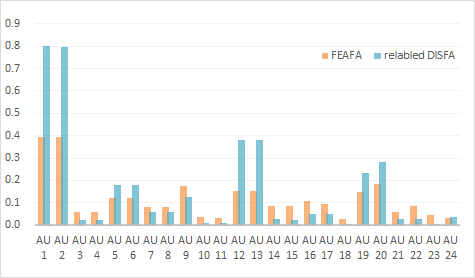}
	\caption{Comparing the occurrence of AUs in FEAFA and relabeled DISFA.}
	\label{fig:occur_compare}
\end{figure}

The occurrence of each AU in FEAFA and the relabeled DISFA is shown visually in Fig \ref{fig:occur_compare}. 
We found that the frequency of common AUs (e.g., AU1/2, AU12/13, and AU19/20) and rare AUs (e.g., AU10/11, AU18, and AU23) in the spontaneous relabeled DISFA was either rather high or rather low, whereas the frequency of each AU in the posed FEAFA was much more moderate.

\begin{figure}[t]
	\centering
	\includegraphics[width=0.95\columnwidth]{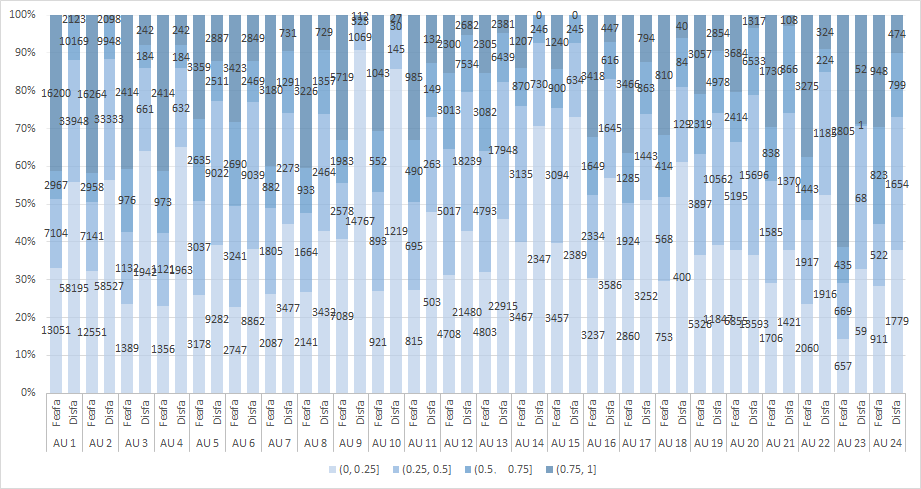}
	\caption{Comparing the distribution of AUs in the range of (0,1] divided into five intervals on FEAFA and relabeled DISFA.}
	\label{fig:distrib}
\end{figure}

The comparison of AU distributions in the range of (0,1] on FEAFA and relabeled DISFA is shown in Fig \ref{fig:distrib}. 
The AUs in FEAFA were more evenly distributed in each interval, and mainly located in the head and tail section. 
The distribution of AUs in the relabeled DISFA was much more extreme, and was largely concentrated in the low value range; some AUs never even appear in the range (0.75,1], such as AU14 and AU15 (i.e., Left/Right Lip Corner Stretch). 
Although the distribution condition of AUs on either posed FEAFA or spontaneous relabeled DISFA mainly depended on the action instruction or the manner of induction, the comparison result can still explain the difference in the numerical distribution between them, to a certain extent.

To compare the difference of AU correlations within FEAFA and relabeled DISFA, as correlation presents both the strength and direction of the association between variables \cite{schober2018correlation}, we used the Pearson correlation coefficient to calculate the linear correlation between each pair of AUs. 
A Pearson correlation is a measure of a linear association between two normally distributed random variables.

By calculating the following Pearson correlation coefficient, we measured the strength between each pair of AUs and their relationships in FEAFA and the relabeled DISFA:
\begin{equation}
	\label{eq:Pearson}
	r=\frac{n(\sum xy)-(\sum x)(\sum y)}{\sqrt{[n\sum x^2-(\sum x)^2][n\sum y^2-(\sum y)^2]}}
\end{equation}
where $x$ and $y$ are the two target variables, and n is the number of pairs of scores. 
The value of $r$ ranges from -1 to 1, where $r=-1$ represents a perfect negative relationship between $x$ and $y$, $r=0$ represents no relationship and $r=1$ represents perfect positive relationship.

\begin{figure}[h]
	\subfigure[AU correlations on FEAFA]{
		\begin{minipage}[t]{0.95\linewidth}
			\centering
			\includegraphics[width=3.4in]{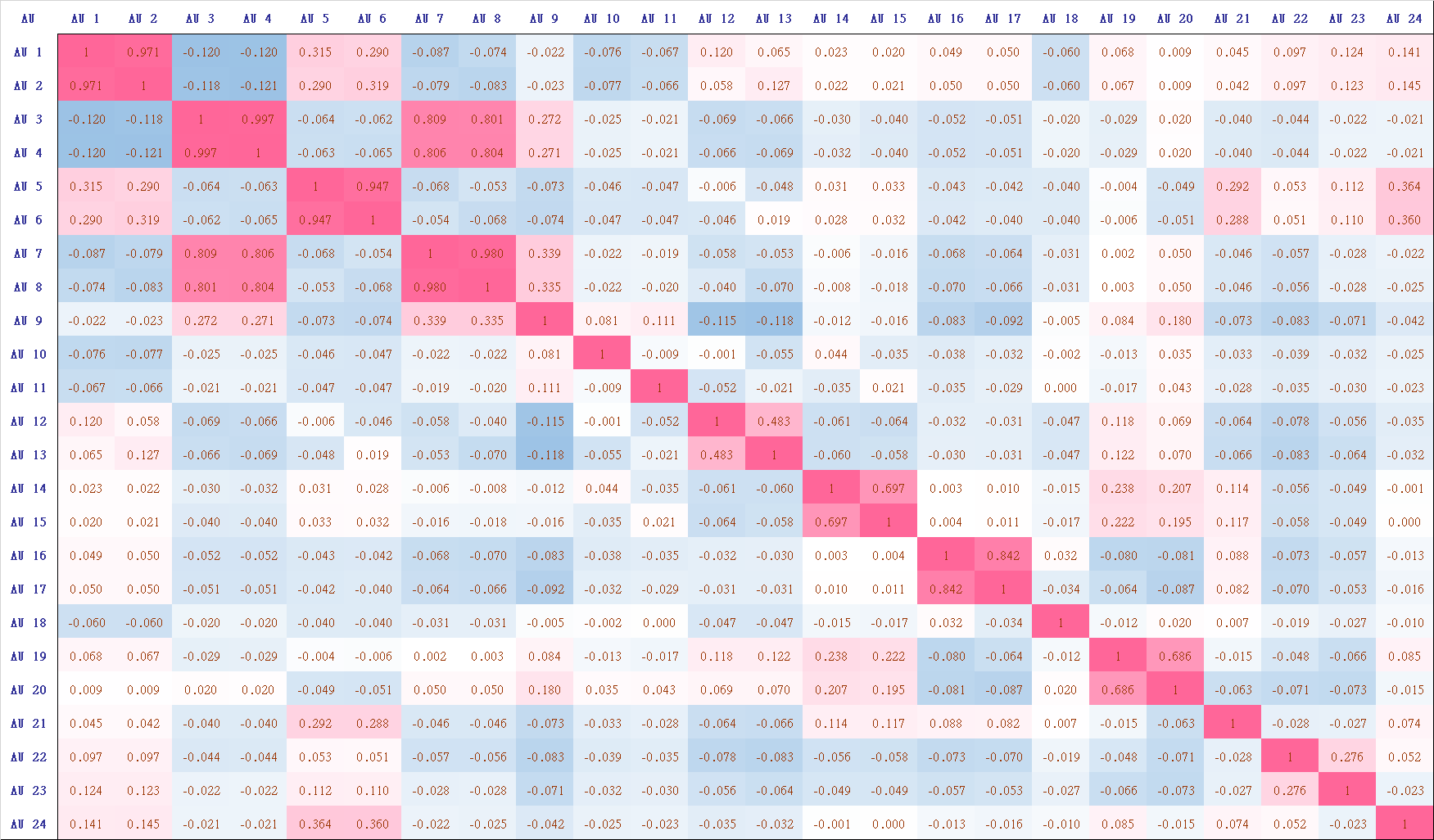}
		\end{minipage}
	}
	\quad
	\subfigure[AU correlations on relabeled DISFA]{
		\begin{minipage}[t]{0.95\linewidth}
			\centering
			\includegraphics[width=3.4in]{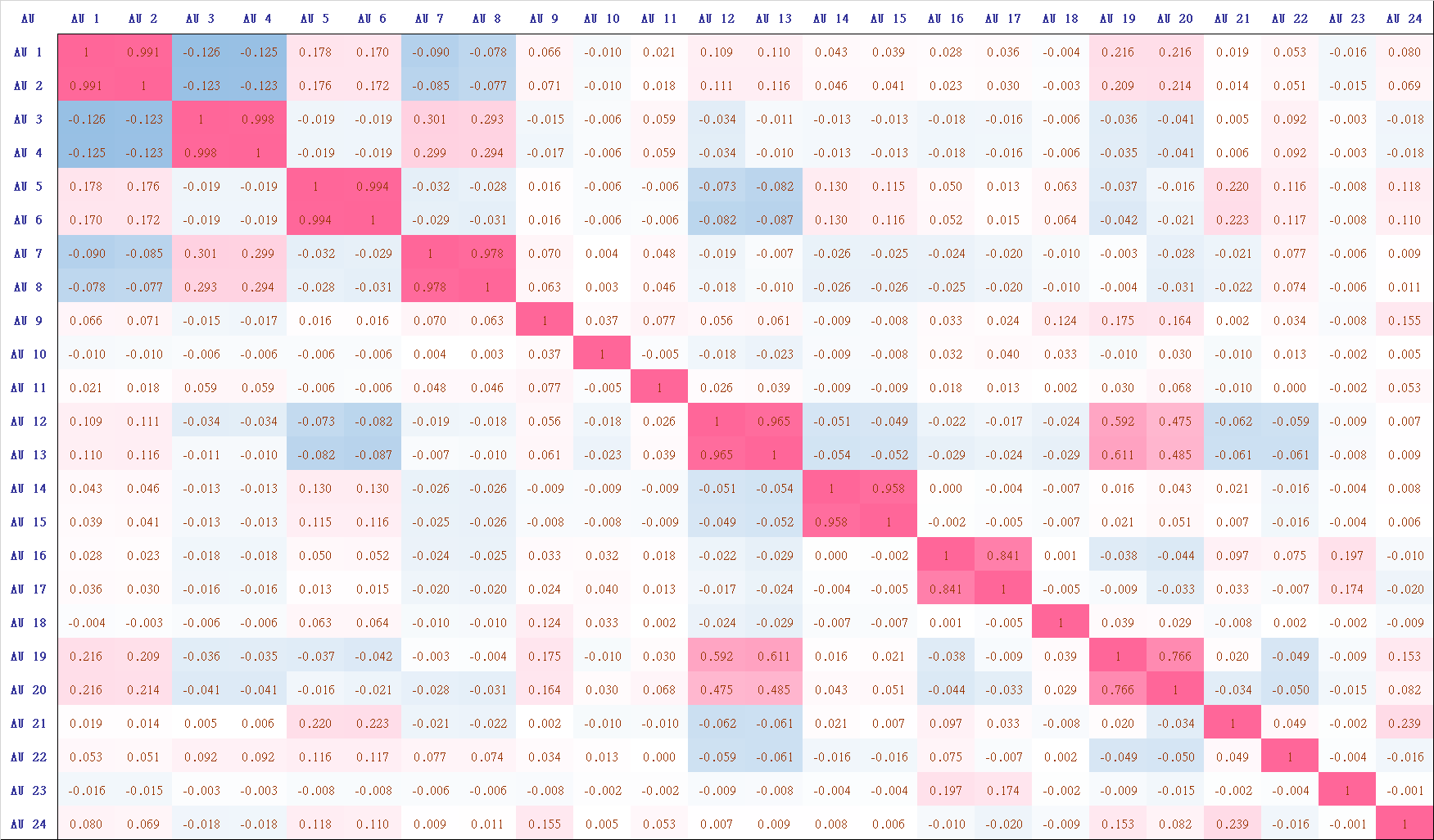}
		\end{minipage}
	}
	\caption{Comparing the correlations between AUs on the posed FEAFA and spontaneous relabeled DISFA: (a) posed AU correlations on FEAFA; (b) spontaneous AU correlations on the relabeled DISFA.}
	\label{fig:corr_compare}
\end{figure}

The calculated Pearson correlation coefficients between AUs on FEAFA and relabeled DISFA are shown in Fig \ref{fig:corr_compare}, from which we can see that for both FEAFA and the relabeled DISFA overall, the asymmetrical AU pairs of AU 1/2, 3/4, 5/6, 7/8, 14/15, 16/17, and 19/20 had strong positive relationships, and the AU pairs of AU 1-2/3-4 and 1-2/7-8 had very weak negative relationships. 
By matching the coefficient value to the definitions of each pair of AUs, we found that were explainable, as we typically close our eyes or lower brows on both sides, and the acts of eye closing and lid raising seldom occur simultaneously.
Additionally, for the spontaneous facial expressions, AU pairs of AU 3-4/7-8 (i.e., Lid Raise/Brow Raise) and 12-13/19-20 (i.e., Lip Corner Pull/Upper Lip Raise or Lower Lip Depress) had some correlation, as people often naturally open their eyes widely with eyebrows raising and smile with teeth visible.

\section{Baseline Evaluation on FEAFA+}
\label{sec:baseline}

\subsection{Image Pre-processing}

We converted the RGB facial images to grayscale images and used the Dlib library to perform face feature point detection, alignment, and cropping on the gray images. 
As the ExpreQuantTool provides functionality for generating AU intensity values by interpolating between keyframes, some AU intensities may have several digits after the decimal point. Thus we rounded all the intensity values to three decimal places. 
Note that we filtered out images whose faces could not be detected at the beginning.

\subsection{Feature Extraction}

After randomly flipping and cropping the images to the size of 224 × 224, we extracted the facial features using VGGNet \cite{Simonyan15} and ResNet \cite{he2016deep}.
VGGNet was proposed to investigate the effect of the convolutional network depth on its performance.
By repeatedly stacking a 3 × 3 convolution kernel and 2 × 2 max pooling layer, it successfully constructs a convolution neural network with 16 to 19 layers.
We used VGG16, which is a 16-layer network composed of 13 convolution layers and three fully connected layers whose  activation functions are all ReLU functions. 
ResNet was proposed to overcome the degradation problem in which the network performance degrades obviously as network depth increases. We adopted ResNet18 with 17 convolution layers and one fully connected layer.
Then 1,000-dimensional deep features could be extracted from the last fully connected layer of these two networks. 

\subsection{Intensity Regression}

Many current approaches use CNNs to recognize a full face image and we also adopted a similar method to estimate AU intensities jointly.

The commonly used evaluation indices of regression tasks are explained variance score, mean absolute error (MAE), mean squared error (MSE), and R squared. We used MSE to evaluate the performance of AU intensity estimation and also used it as the loss function with the aim to minimize the following MSE sum loss:
\begin{equation}
	\label{eq:loss}
	L=\frac{1}{N}\sum_{i=1}^{N}\sum_{j=1}^{24}\Vert\textbf{x}_{ij}-\hat{\textbf{x}}_{ij}\Vert^{2},
\end{equation}
where $L$ is the sum loss of the training samples, and $x_{ij}$ and $\hat{x}_{ij}$ are the predicted value and ground truth value of AU-j respectively. 

We implemented stochastic gradient descent as the optimizer and used the StepLR method to adjust the learning rate, thereby decaying the learning rate of each parameter group by 0.1 every 10 epochs.
We provided a baseline performance for AU value regression conducted on FEAFA+ using the PyTorch framework and performed the training process on an NVIDIA GeForce RTX 3090 GPU machine. 
A three-fold cross validation were adopted to divide the training and validation data based on video sequences.
We used the models subpackage of PyTorch directly without pretraining and simply made changes on the fully connected layers of the networks.
The regression results of the two networks are listed in Table \ref{tab:reg}.
\begin{table}[h]
	\centering
	\caption{Regression results for two CNNs on the FEAFA+ dataset.}
	\label{tab:reg}
	\footnotesize
	\begin{tabular}{c|c|c}
		\hline
		Dataset                       & baseline & MSE    \\ \hline
		\multirow{2}{*}{relabeled DISFA} & resnet18 & \textbf{0.3209} \\ \cline{2-3} 
		& vgg16    & 0.3285 \\ \hline
		\multirow{2}{*}{FEAFA}    & resnet18 & 0.307  \\ \cline{2-3} 
		& vgg16    & \textbf{0.2941} \\ \hline
	\end{tabular}
\end{table}
However, as using CNNs to recognize the full image and not taking individual region's labels into account may incorrectly connect the image contexts, for example, connecting eye-area-related AUs with mouth-area-related features \cite{ma2019r}, more effort is required to learn accurate regional features of AUs, and expert prior knowledge may need to be adopted in the future.


\section{Conclusions}
\label{sec:conclusions}

In this paper, we presented a new well-annotated FACS-labeled dataset of posed and spontaneous facial expressions recorded in a laboratory environment and real-world conditions, respectively. All the images were manually annotated on 0--1 floating-point intensities of 24 redefined AUs. We provided a broad comparison of the numerical analysis of the posed and spontaneous subsets. We also presented baseline regression results for two CNNs. 
With FEAFA+, researchers can obtain smooth and continuous AU values of given facial images or videos using AU intensity estimation algorithms. We hope that more investigations and applications will be conducted in the future with the help of FEAFA+, and we will keep updating the publicly available dataset and conducting research in relevant fields.


\section*{Acknowledgements}
	This work is supported by National Key R\&D Program of China under contract No. 2017YFB1002203, National Natural Science Foundation of China (NSFC, Grant Nos. 62032022, 61972375, 61929104, 61671426), the Beijing Natural Science Foundation (Grant No. 4182071). 

\bibliographystyle{elsarticle-num}
\bibliography{feafa+}

\end{document}